\documentclass{article}

\PassOptionsToPackage{numbers,sort&compress}{natbib}
\usepackage[preprint]{neurips_2026}
\workshoptitle{NeurIPS 2026 Workshop} 
\usepackage{float}
\usepackage[utf8]{inputenc}
\usepackage[T1]{fontenc}
\usepackage[hidelinks]{hyperref}
\usepackage{url}
\usepackage{booktabs}
\usepackage{amsmath,amsfonts}
\usepackage{microtype}
\usepackage{xcolor}
\usepackage{graphicx}
\usepackage{caption}
\usepackage{ragged2e}
\usepackage{tikz}
\usetikzlibrary{arrows.meta,positioning}

\renewcommand{\bibfont}{\small} 

\providecommand{\name}{\bfseries}
\providecommand{\addr}{\normalfont}

\title{FlowTSFM: Turning Encoder Depth \\ into Quantile Transport}

\author{%
    \parbox{\textwidth}{\centering
        \name
        Bahaeddine Abdessalem$^{1,3}$ \quad
        Shifeng Xie$^{1,2}$ \quad
        Zehao Xiao$^{1}$ \\[0.4em]
        \name
        Youssef Attia El Hili$^{1,4}$ \quad
        Ambroise Odonnat$^{1,5}$ \quad
        \name
        Jianfeng Zhang$^{6}$ \quad
        Lujia Pan$^{6}$ \\ \quad
        Keli Zhang$^{1}$ \quad
        Malik Tiomoko$^{1}$ \\[0.8em]
        \addr
        $^{1}$Huawei Noah's Ark Lab, Paris, France \\
        \addr
        $^{2}$LIPADE, Universit\'e Paris Cit\'e, Paris, France \\
        \addr
        $^{3}$\'Ecole Polytechnique, France \\
        \addr
        $^{4}$Centre de Recherche en Informatique,
        Mines Paris, PSL University, France \\
        \addr
        $^{5}$IRISA, Universit\'e Rennes 2, Inria, France \\
        \addr
        $^{6}$Huawei Noah's Ark Lab, Shenzhen, China \\
    }%
}
\begin{document}

\maketitle
\begin{abstract}

 Encoder-based time series foundation models (TSFMs) typically rely on deep stacks of independently parameterized Transformer layers, where only the final forecast is supervised and intermediate representations have no explicit predictive role. We introduce FlowTSFM, an encoder architecture that interprets depth as a recurrent transport process: a single Transformer block is iteratively applied with shared parameters, while a quantile-flow objective supervises intermediate states along a prescribed trajectory from a prior distribution toward the final forecast. The objective combines pinball forecasting loss with path-level position matching. With only 38.8M parameters, FlowTSFM achieves competitive performance on GIFT-Eval and TIME, remaining within 1.8–4.6\% MASE of stronger baselines while using approximately 3× fewer parameters than a 12-layer Chronos-2 model (119.5M). Beyond accuracy, we introduce CosMean, a scale-free diagnostic measuring whether recurrent updates consistently align toward the final prediction. Under a matched intermediate-state probing protocol, FlowTSFM achieves a CosMean score of 0.919 compared with 0.350 for Chronos-2, suggesting that recurrent parameter sharing combined with path supervision is associated with substantially more structured predictive trajectories at a favorable accuracy-efficiency trade-off.

\end{abstract}

\section{Introduction}

Time series forecasting is an important task across several domains, such as finance \citep{sezer2020financial}, healthcare \citep{johnson2023mimic}, cloud operations \citep{joosen2023serverless}, and industrial monitoring \citep{yan2024industrial}.  Forecasts
influence consequential decisions (support planning, resource
allocation, risk management) yet real-world signals differ widely in
scale, sampling rate, noise, and temporal structure
\citep{podest2026tirex2}, motivating general-purpose probabilistic
forecasters, often built on patch-based representations
\citep{nie2023patchtst}, that transfer across datasets and domains
rather than requiring a separate model per series \citep{ansari2024chronos}. 

Recent time-series foundation models achieve strong forecasting performance through encoder-based, decoder-based, and recurrent architectures \citep{ansari2024chronos,das2024timesfm,woo2024unified,podest2026tirex2}. However, in encoder-based models, depth remains largely an internal computation: intermediate states have no explicit forecasting role, while supervision ultimately targets the resulting forecast. Generative flow matching offers a contrasting view in which intermediate computation describes transport along a path between distributions \citep{lipman2023flowmatching}, but this idea has primarily been developed for sample-space generative modeling. It therefore remains unclear whether encoder depth in probabilistic forecasting can instead be organized as a structured transport process over predictive quantiles.

We introduce FlowTSFM, a forecasting framework that interprets encoder depth as quantile transport. FlowTSFM replaces the multiple independent encoder layers with a single shared block applied recurrently through an ODE-like residual
update \citep{chen2018neuralode}, in the spirit of recent work on
weight-tied recursive computation \citep{jolicoeur2025trm}. We further introduce a quantile-flow objective that supervises intermediate predictions along a prescribed trajectory toward the terminal forecast. Unlike generative flow matching, our formulation does not learn a sample-space velocity field; the recurrence evolves in latent space while its decoded predictive quantiles are directly constrained along depth.

\begin{figure}[t]
  \centering
  \includegraphics[width=\linewidth]{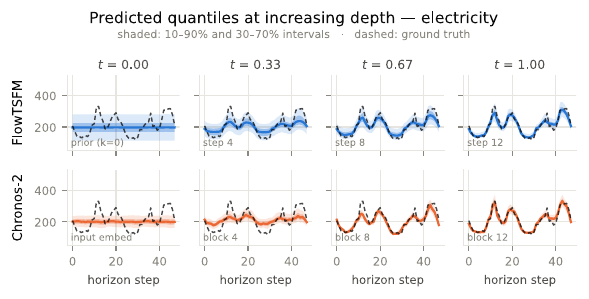}
  \caption{\textbf{Depth-wise quantile evolution on a held-out Electricity-H series.} Columns correspond to increasing normalized depth $t=\frac kK$. In each panel, the solid line is the median forecast, the darker and lighter bands are the $30$--$70\%$ and $10$--$90\%$ predictive intervals, and the dashed line is the ground truth. FlowTSFM (top row) already forms a coherent forecast at early exits and refines it smoothly across depth. Chronos-2 (bottom row), probed with the same frozen output head, shows weaker early forecast structure and larger late-stage corrections.}
  \label{fig:transport}
\end{figure}


 Empirically, FlowTSFM (38.8 parameters) comes within 1.8--4.6\% MASE
 of the strongest baselines on GIFT-Eval \citep{aksu2024gifteval} and
 TIME \citep{qiao2026time}, while using roughly $3\times$ fewer
 parameters than a 12-layer Chronos-2 (119.5M). Under a matched
 trajectory protocol, FlowTSFM reaches a mean
 cosine alignment (CosMean) of 0.919 toward its terminal forecast,
 compared to 0.350 for an equivalently probed Chronos-2 (Figure~\ref{fig:transport}).
 Together, these results indicate that intermediate encoder computation
 can be given an explicit geometric role at a favorable
 accuracy--efficiency trade-off, rather than as a strict improvement
 over larger, non-recurrent encoders. Concretely, our contributions are:
 \begin{itemize}
   \item[(i)] a recurrent encoder architecture that replaces independent
   Transformer layers with a single weight-tied block, cutting parameter
   count by roughly $3\times$ relative to a comparable 12-layer model;
   \item[(ii)] a quantile-flow training objective that supervises
   intermediate decoder exits along a prescribed source-to-forecast
   trajectory, giving encoder depth an explicit predictive role;
   \item[(iii)] We employ CosMean\cite{liu2023rectified}, a scale-free diagnostic  for whether recurrent
   updates advance toward the terminal forecast, and a matched
   trajectory protocol for applying it across architecturally different
   models.
 \end{itemize}

\section{Methodology}
\paragraph{Notation.}
We write $[n]=\{1,\ldots,n\}$. The normalized input is $x$ and the
observation mask is $m$; the encoder $E$ maps the pair $(x,m)$ to the
initial latent state $h^0=E(x,m)\in\mathbb R^{T\times d_{\text{model}}}$, where $d_{\text{model}}$ is
the latent space dimension. Let $\mathcal M\subseteq[T]$ be the set of forecast
positions scored by the loss, and $y_s$ the target at position
$s\in\mathcal M$.

We discretize depth into $K$ steps: for $k\in\{0,\ldots,K\}$ we set
$t_k=\frac kK$, with step size $\Delta t=\frac 1K$. Each $h^k\in\mathbb R^{T\times
d_{\text{model}}}$ is the latent state at depth $k$, and the decoder $D_\phi$ maps it to
the decoded field $q^k=D_\phi(h^k)=(q^k_{s,j})\in\mathbb R^{T\times Q}$,
whose entry $q^k_{s,j}$ is the prediction at position $s$ for quantile
level $\tau_j$, $j\in[Q]$. We call each selected depth $k$ an
\emph{exit}.

Finally, $r=(r_j)_{j\in[Q]}$ is the analytic \emph{source} attached to
the quantile levels, defined by $r_j=\operatorname{arcsinh}
(\Phi^{-1}(\tau_j))$ and broadcast across all forecast positions. This analytic source represents the decoding of the first hidden layer that we obtain directly after the embedding of the time series. The analytic source represents the default quantile predictions of the model, which in this case we will use the quantiles of a normal distribution. Therefore we use $r_j=\operatorname{arcsinh}
(\Phi^{-1}(\tau_j))$ rather than$r_j= \Phi^{-1}(\tau_j)$ because we further use $\operatorname{arcsinh}$ \cite{ansari2025chronos2} normalization.

\paragraph{Recurrent encoder architecture}
\label{sec:architecture}

A conventional $K$-layer encoder computes $h^{k+1}=B_{\theta_k}(h^k)$ with independently learned $\theta_k$. FlowTSFM instead reuses one shared velocity field $F_\theta$ at every depth:
\begin{equation}
 h^{k+1}=h^k+\Delta t\,F_\theta\!\left(h^k,e_t(t_k),e_{\text{depth}}(K)\right),
 \qquad k=0,\ldots,K-1.
 \label{eq:euler}
\end{equation}
\par\begingroup
\ifdefined\nolinenumbers\nolinenumbers\fi
\noindent
\begin{minipage}[t]{0.47\linewidth}
  \vspace{0pt}
  \justifying
  \ifdefined\internallinenumbers\internallinenumbers\fi
Here $e_t(t_k)$ encodes the normalized depth and $e_{\text{depth}}(K)$ encodes the total step budget, allowing the same recurrent block to adapt to different unroll lengths. Both $F_\theta$ and the output chart $D_\phi$ are shared across depth. This update resembles an ODE integration in the latent space\cite{chen2018neuralode}
 
\paragraph{Quantile-flow objective}
\label{sec:objective}

Let $\ell_{\mathrm{pin}}^k$ be the mean \emph{pinball loss} at exit $k$, i.e., the standard proper loss for quantile forecasting, and let
$\pi^k=(1-t_k)r+t_k\operatorname{sg}(q^K)$ be the path target interpolating between the source and the terminal prediction where $\operatorname{sg}$ denotes the stop-gradient operator, which leaves its input unchanged in the forward pass but blocks gradient propagation during backpropagation. During training, we supervise a fixed number of intermediate exits: at each step, a fixed-size subset $\mathcal E'\subseteq\{1,\ldots,K-1\}$ is sampled uniformly. For these sampled exits, we minimize
\end{minipage}\hfill
\begin{minipage}[t]{0.49\linewidth}
  \vspace{0pt}
  \centering
  \resizebox{\linewidth}{!}{
\begin{tikzpicture}[
  x=1mm,y=1mm,
  font=\sffamily\fontsize{8}{9}\selectfont,
  box/.style={draw=black!55,line width=.45pt,rounded corners=1pt,
    align=center,minimum height=7.4mm,inner xsep=1mm,inner ysep=.7mm},
  io/.style={box,fill=blue!7,text width=13mm},
  layer/.style={box,fill=orange!10,minimum width=7mm},
  head/.style={box,fill=green!9,text width=14mm},
  arr/.style={-{Latex[length=1.3mm,width=.95mm]},line width=.55pt,draw=black!75},
  panel/.style={draw=black!20,fill=black!1,rounded corners=2pt,line width=.45pt},
  heading/.style={anchor=west,font=\sffamily\fontsize{8}{9}\selectfont\bfseries},
  note/.style={font=\sffamily\fontsize{8}{9}\selectfont,text=black!70,align=center}
]
\path[use as bounding box] (0,0) rectangle (77,53);

\draw[panel] (0,31) rectangle (77,53);
\node[heading] at (2.5,49) {(a) Conventional encoder};
\node[io] (ea) at (9.5,40.5) {patch\\embedder};
\node[layer] (b1) at (22.5,40.5) {$B_1$};
\node[layer] (b2) at (31.5,40.5) {$B_2$};
\node[inner sep=0pt] (dots) at (38.5,40.5) {$\cdots$};
\node[layer] (bk) at (45.5,40.5) {$B_K$};
\node[head] (da) at (58,40.5) {quantile\\head};
\node[inner sep=0pt] (qa) at (71.5,40.5) {$q^K$};
\draw[arr] (ea.east)--(b1.west);
\draw[arr] (b1.east)--(b2.west);
\draw[draw=black!75,line width=.55pt] (b2.east)--(36.1,40.5);
\draw[arr] (40.9,40.5)--(bk.west);
\draw[arr] (bk.east)--(da.west);
\draw[arr] (da.east)--(qa.west);

\draw[panel] (0,0) rectangle (77,28.5);
\node[heading] at (2.5,24.5) {(b) FlowTSFM};
\node[io] (eb) at (10.5,14.0) {patch\\embedder};
\node[layer,text width=10mm] (fb) at (31.5,14.0) {$F_\theta$};
\node[head] (db) at (52.5,14.0) {quantile\\head};
\node[inner sep=0pt, anchor=west] (qb) at (62.6,14.0) {$q^0,\ldots,q^K$};
\draw[arr] (eb.east)--(fb.west);
\draw[arr] (fb.east)--(db.west);
\draw[arr] (db.east)--(qb.west);
\draw[arr]
  ([xshift=2.0mm,yshift=-0.2mm]fb.south)
    .. controls +(5.5mm,-7.5mm) and +(-5.5mm,-7.5mm) ..
  ([xshift=-2.0mm,yshift=-0.2mm]fb.south);
\node[note] at (31.5,2.9) {$h^k$};
\end{tikzpicture}
}
  \begingroup
    \captionsetup{font=small,justification=justified,singlelinecheck=false,hypcap=false}
    \captionof{figure}{Conventional depth (top) uses independent blocks $B_1,\ldots,B_K$. FlowTSFM (bottom) replaces them with one recurrent block applied repeatedly, producing intermediate forecasts $q^0,\ldots,q^K$. See Appendix~\ref{app:architecture} for implementation details.}
    \label{fig:architecture}
  \endgroup
\end{minipage}
\par\endgroup\medskip
\begin{equation}
 \mathcal L
 =\ell_{\mathrm{pin}}^K
 +\sum_{k\in\mathcal E'}
 \left[
   \lambda_{\mathrm{ds}}t_k\ell_{\mathrm{pin}}^k
   +\lambda_{\mathrm{qf}}\left\|q^k-\pi^k\right\|_{\mathcal M}^2
 \right]
 +\lambda_0\left\|q^0-r\right\|_{\mathcal M}^2 .
 \label{eq:objective}
\end{equation}
Here $\|\cdot\|_{\mathcal M}^2$ is the mean squared deviation over scored positions and quantile levels. The first term $\ell_{\mathrm{pin}}^K$ optimizes the terminal forecast $q^K$; the depth-scaled intermediate pinball term keeps early exits predictive; the path-matching term makes intermediate decoded quantiles follow a prescribed source-to-endpoint trajectory; and the last term anchors the learned initial exit $q^0$ to the source $r$. The stop-gradient ensures that this auxiliary path supervision shapes the trajectory without directly moving the terminal target. Exact definitions are deferred to Appendix~\ref{app:architecture}.
\section{Results}
\label{sec:results}
\paragraph{Experimental setting.}
The accuracy test makes use of an 8192-point context, a patch size of 16, $K=10$, 99 quantiles, and a 50/50 combination of domain-balanced GIFT-Eval-Pretrain data and kernel-generated synthetic data. It is trained for 100,000 AdamW steps and is assessed via the Seasonal-Naive-normalized geometric aggregations as specified by GIFT-Eval and TIME. For trajectory evaluation, we use the released 12-layer Chronos-2.

\begin{table}[!htbp]
\caption{Seasonal-Naive-normalized forecasting accuracy on the two standard benchmarks. GIFT-Eval contains 97 configurations and TIME contains 98 tasks. Lower is better.}
  \label{tab:main-results}
  \centering
  \small
  \setlength{\tabcolsep}{4pt}
  \begin{tabular}{lrrrrr}
    \toprule
    Model & GIFT MASE $\downarrow$ & GIFT CRPS $\downarrow$ & TIME MASE $\downarrow$ & TIME CRPS $\downarrow$  & Paramaters \\
    \midrule
    FlowTSFM (base) & 0.729 & 0.500 & 0.694 & 0.580 & 38.8M \\
    Sundial (base) & 0.750 & 0.559 & 0.758 & 0.663  & 128.3M\\
    Tirex  & 0.716 & 0.488 & 0.683 & 0.573 & 35.3M \\
    Moirai-2  & 0.728 & 0.516 & 0.703 & 0.588  & 11.4M\\
    Timesfm 2.0 & 0.758 & 0.550 & 0.718 & 0.620 & 498.8M\\
    Chronos-2 & \textbf{0.697} & \textbf{0.485} & \textbf{0.662} & \textbf{0.556} & 119.5M\\
    \bottomrule
  \end{tabular}
\end{table}

\paragraph{Metrics.}
Following directional diagnostics for rectified and self-consistent flows
\citep{liu2023rectified,han2026selfconsistent}, we use only mean cosine
alignment between each decoded update and the remaining displacement to $q^K$. Intuitively, this metric is high when the forecast trajectory advances toward its terminal prediction rather than wandering or reversing:
\begin{equation}
\operatorname{CosMean}(q)
=
\frac{1}{K-1}\sum_{k=0}^{K-2}
\frac{
\left\langle q^{k+1}-q^k,\,q^K-q^k\right\rangle
}{
\left\|q^{k+1}-q^k\right\|_2\left\|q^K-q^k\right\|_2
}.
\label{eq:meancos}
\end{equation}
Inner products flatten $(s,j)\in\mathcal M\times[Q]$. The final update is
omitted because its cosine is one by construction. Hence
$\operatorname{CosMean}(q)\in[-1,1]$, with larger values indicating more
endpoint-directed transport.
\begin{table}[t]
  \centering
  \caption{Mean cosine alignment between local trajectory velocities and the
  endpoint chord on GIFT-Eval. Dataset-macro averages weight each of the 55
  datasets equally, whereas path-weighted averages weight every valid test
  trajectory equally. Higher is better.}
  \label{tab:gifteval-meancos}
  \begin{tabular}{lcc}
    \toprule
    Model & Dataset macro $\uparrow$ & Path weighted $\uparrow$ \\
    \midrule
    FlowTSFM  & \textbf{0.919} & \textbf{0.933} \\
    Chronos-2 & 0.350          & 0.317          \\
    \bottomrule
  \end{tabular}
\end{table}

\section{Discussion and conclusion}

FlowTSFM makes two linked changes to the encoder-only architecture: it replaces the K separate layers with K instances of an ODE-like field in latent space, and it substitutes terminal-only training with a quantile-flow objective which gives the recurrent interior a well-defined purpose by constraining it. The reduction in the number of parameters is a direct result of weight tying. The geometric findings indicate that the objective has successfully assigned a coherent role to the repeated computation and produced more directionally aligned quantile transport.

The evidence remains preliminary: accuracy and geometry use different single-run checkpoints; Chronos-2's early states were not trained or supervised in the same way as in our current setting. Furthermore, ablation studies for the new objective, along with tests with different values of $K$, are still pending. The narrower result is that decoded path supervision yields a much more structured recurrent trajectory at comparable terminal quality.

\clearpage
\nocite{*}

\bibliographystyle{plainnat}
\bibliography{neurips_workshop}
\clearpage
\appendix
\raggedbottom
\section{Related work}
Generative flow matching \cite{lipman2023flowmatching} learns a time-dependent vector field that transports a simple source distribution toward a data distribution along a prescribed probability path. Training samples a time and a state on this path and regresses the model toward the corresponding conditional velocity, avoiding likelihood evaluation and full ODE simulation during training \citep{lipman2023flowmatching,chen2018neuralode}. Recent work studies shortcut parameterizations and the compression of such transport paths into very few, or even one, generation steps \citep{frans2024shortcut,han2026wflow}. In parallel, time-series foundation models pretrain probabilistic forecasters on heterogeneous corpora: patch-based architectures underpin Chronos, TimesFM, Moirai, and Chronos-2 \citep{ansari2024chronos,das2024timesfm,woo2024unified,ansari2025chronos2}, while TiRex-2 demonstrates the continued relevance of recurrent designs \citep{podest2026tirex2}. 
\section{Detailed architecture and training configuration}
\label{app:architecture}
\label{app:details}

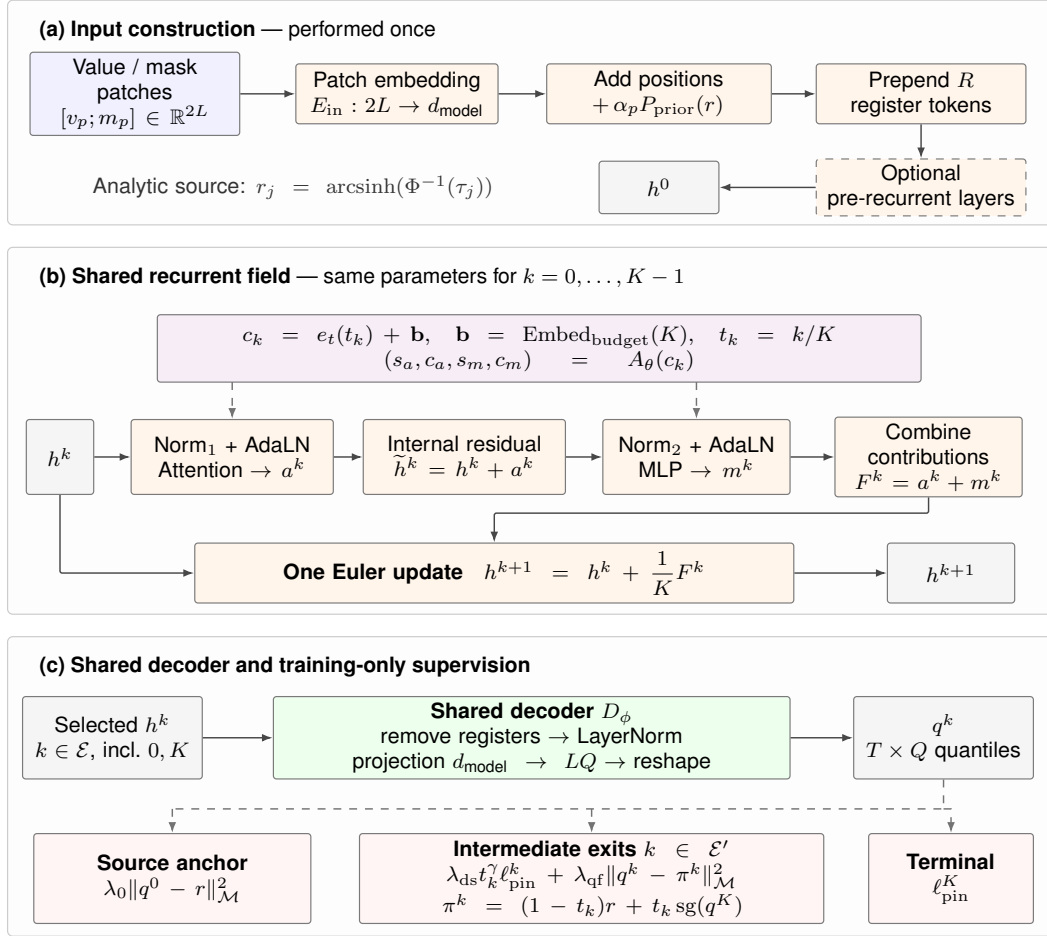
\begin{figure}[H]
  \centering
  \resizebox{\linewidth}{!}{
\begin{tikzpicture}[
  x=1mm,y=1mm,
  font=\sffamily\fontsize{8}{9.5}\selectfont,
  box/.style={draw=black!52,line width=.45pt,rounded corners=1.2pt,
    align=center,inner xsep=1mm,inner ysep=.8mm,minimum height=8mm},
  input/.style={box,fill=blue!6},
  operation/.style={box,fill=orange!8},
  state/.style={box,fill=black!4},
  decoder/.style={box,fill=green!7},
  control/.style={box,fill=violet!7},
  objective/.style={box,fill=red!4},
  arrow/.style={-{Latex[length=1.45mm,width=1.05mm]},
    line width=.55pt,draw=black!72},
  auxiliary/.style={arrow,dashed,draw=black!55},
  panel/.style={draw=black!20,fill=black!0.6,line width=.45pt,
    rounded corners=2pt},
  heading/.style={anchor=west,font=\sffamily\fontsize{8}{10}\selectfont\bfseries},
  note/.style={font=\sffamily\fontsize{8}{9.5}\selectfont,text=black!80,align=center}
]

\draw[panel] (0,95) rectangle (140,125);
\node[heading] at (3,121) {(a) Input construction \normalfont\sffamily--- performed once};
\node[input,text width=26mm] (patches) at (17,112.5)
  {Value / mask patches\\$[v_p;m_p]\in\mathbb R^{2L}$};
\node[operation,text width=25mm] (embedding) at (52,112.5)
  {Patch embedding\\$E_{\rm in}:2L\to d_{\text{model}}$};
\node[operation,text width=28mm] (prior) at (87,112.5)
  {Add positions\\$+\,\alpha_pP_{\rm prior}(r)$};
\node[operation,text width=26mm] (registers) at (122,112.5)
  {Prepend $R$\\register tokens};
\draw[arrow] (patches.east)--(embedding.west);
\draw[arrow] (embedding.east)--(prior.west);
\draw[arrow] (prior.east)--(registers.west);

\node[operation,dashed,text width=26mm,minimum height=7mm] (prelayers) at (122,100)
  {Optional\\pre-recurrent layers};
\node[state,minimum width=16mm,minimum height=7mm] (initial) at (87,100) {$h^0$};
\draw[arrow] (registers.south)--(prelayers.north);
\draw[arrow] (prelayers.west)--(initial.east);
\node[note,text width=71mm] at (38,100)
  {Analytic source: $r_j=\operatorname{arcsinh}(\Phi^{-1}(\tau_j))$};

\draw[panel] (0,43) rectangle (140,92);
\node[heading] at (3,88) {(b) Shared recurrent field \normalfont\sffamily--- same parameters for $k=0,\ldots,K-1$};
\node[control,text width=100mm,minimum height=9mm] (condition) at (71,78.5)
  {$c_k=e_t(t_k)+\mathbf b,\quad \mathbf b=\operatorname{Embed}_{\mathrm{budget}}(K),\quad t_k=k/K$\\
   $(s_a,c_a,s_m,c_m)=A_\theta(c_k)$};

\node[state,minimum width=9mm,minimum height=10mm] (current) at (7,64) {$h^k$};
\node[operation,text width=25mm,minimum height=10mm] (attention) at (30,64)
  {Norm$_1$ + AdaLN\\Attention $\to a^k$};
\node[operation,text width=25mm,minimum height=10mm] (residual) at (61,64)
  {Internal residual\\$\widetilde h^k=h^k+a^k$};
\node[operation,text width=23mm,minimum height=10mm] (mlp) at (92,64)
  {Norm$_2$ + AdaLN\\MLP $\to m^k$};
\node[operation,text width=23mm,minimum height=10mm] (field) at (123,64)
  {Combine contributions\\$F^k=a^k+m^k$};
\draw[arrow] (current.east)--(attention.west);
\draw[arrow] (attention.east)--(residual.west);
\draw[arrow] (residual.east)--(mlp.west);
\draw[arrow] (mlp.east)--(field.west);
\draw[auxiliary] (30,74)--(attention.north);
\draw[auxiliary] (92,74)--(mlp.north);

\node[operation,text width=78mm,minimum height=8mm] (euler) at (65,48.5)
  {\textbf{One Euler update}\quad $h^{k+1}=h^k+\dfrac{1}{K}F^k$};
\node[state,minimum width=17mm,minimum height=8mm] (next) at (126,48.5) {$h^{k+1}$};
\draw[arrow,rounded corners=.8pt] (current.south)--(7,48.5)--(euler.west);
\draw[arrow,rounded corners=.8pt] (field.south)--(123,56.5)--(65,56.5)--(euler.north);
\draw[arrow] (euler.east)--(next.west);

\draw[panel] (0,0) rectangle (140,40);
\node[heading] at (3,36) {(c) Shared decoder and training-only supervision};
\node[state,text width=22mm,minimum height=11mm] (selected) at (14,26.5)
  {Selected $h^k$\\$k\in\mathcal E$, incl.\ $0,K$};
\node[decoder,text width=67mm,minimum height=11mm] (readout) at (70,26.5)
  {\textbf{Shared decoder $D_\phi$}\\
   remove registers $\to$ LayerNorm\\
   projection $d_{\text{model}}\to LQ$ $\to$ reshape};
\node[state,text width=22mm,minimum height=11mm] (quantiles) at (125,26.5)
  {$q^k$\\$T\times Q$ quantiles};
\draw[arrow] (selected.east)--(readout.west);
\draw[arrow] (readout.east)--(quantiles.west);

\node[objective,text width=35mm,minimum height=12mm] (anchor) at (22,7.7)
  {\textbf{Source anchor}\\$\lambda_0\|q^0-r\|_{\mathcal M}^2$};
\node[objective,text width=60mm,minimum height=12mm] (interior) at (78,7.7)
  {\textbf{Intermediate exits} $k\in\mathcal E'$\\
   $\lambda_{\rm ds}t_k^\gamma\ell_{\rm pin}^k
    +\lambda_{\rm qf}\|q^k-\pi^k\|_{\mathcal M}^2$\\
   $\pi^k=(1-t_k)r+t_k\operatorname{sg}(q^K)$};
\node[objective,text width=20mm,minimum height=12mm] (terminal) at (126,7.7)
  {\textbf{Terminal}\\$\ell_{\rm pin}^K$};
\draw[dashed,line width=.55pt,draw=black!55] (quantiles.south)--(125,17)--(22,17);
\draw[auxiliary] (22,17)--(anchor.north);
\draw[auxiliary] (78,17)--(interior.north);
\draw[auxiliary] (125,17)-|(terminal.north);
\end{tikzpicture}}
  \caption{Detailed FlowTSFM computation. (a) Input construction runs once; optional pre-recurrent layers are outside the loop. (b) Attention and MLP contributions form one shared field; only the Euler block updates the recurrent state. (c) The shared decoder also reads $h^0$, so $q^0$ is learned and anchored to $r$. Dashed arrows denote conditioning or training-only supervision.}
  \label{fig:detailed-architecture}
\end{figure}

Figure~\ref{fig:detailed-architecture} expands the architecture in Section~\ref{sec:architecture}, separating the one-time input construction, the shared recurrent field, and the decoder with its training objectives.

\clearpage

\paragraph{Notation and preprocessing.}
Let $x_{\rm raw}$ be the raw series and $m$ its binary observation mask. Following the instance-normalization principle of ReVIN \citep{kim2022reversible}, we compute $\mu$ and $\sigma$ from visible context values only and set $x=\operatorname{arcsinh}((x_{\rm raw}-\mu)/\sigma)$; masked values do not contribute to these statistics, and targets $y_s$ use the same coordinates. Writing $[n]=\{1,\ldots,n\}$, the shared decoder produces $q^k=D_\phi(h^k)=(q^k_{s,j})_{s\in[T],j\in[Q]}\in\mathbb R^{T\times Q}$ at every selected exit $k\in\mathcal E\subseteq\{0,\ldots,K\}$, while losses retain only $s\in\mathcal M\subseteq[T]$. The source $r=(r_j)_{j\in[Q]}$, with $r_j=\operatorname{arcsinh}(\Phi^{-1}(\tau_j))$, is broadcast across positions. In particular, $q^0=D_\phi(h^0)$ is learned and is anchored to $r$ by the final term of Eq.~\ref{eq:objective}; it is not defined as $r$. We write $\mathcal E'\subseteq\{1,\ldots,K-1\}$ for the intermediate supervised exits sampled during training; at each training step, a fixed-size subset is drawn uniformly from the available interior exits. Superscripts index recurrent depth, not exponentiation.

\paragraph{Loss definitions.}
For any exit $k$, the compact terms in Eq.~\ref{eq:objective} are
\begin{align*}
 \ell_{\mathrm{pin}}^k
 &=\frac{1}{|\mathcal M|Q}
   \sum_{s\in\mathcal M}\sum_{j=1}^{Q}
   \rho_{\tau_j}(y_s-q^k_{s,j}),
 &\rho_\tau(u)&=\max\{\tau u,(\tau-1)u\},\\
 \|A\|_{\mathcal M}^2
 &=\frac{1}{|\mathcal M|Q}
   \sum_{s\in\mathcal M}\sum_{j=1}^{Q}A_{s,j}^2 .
\end{align*}
Thus $\|q^k-\pi^k\|_{\mathcal M}^2$ is deterministic position matching
to the linear source--endpoint path. Since $t_0=0$, $\pi^0=r$, so the
last term of Eq.~\ref{eq:objective} is exactly the learned exit-$0$
source anchor. Stop-gradient blocks gradients through the terminal
value used to construct $\pi^k$; the recurrent field and decoder remain
shared trainable parameters.

\paragraph{Input and recurrent block.}
Length-$L$ value patches are concatenated with their union-mask patches and mapped from $2L$ to width $D$ by the input patch embedding. Learned positional embeddings are added in the reported configurations. A learned projection $P_{\rm prior}(r)$ of the source vector is added to each patch in proportion to its prediction-mask fraction $\alpha_p$, after which $R$ learned register tokens are prepended. This injection conditions the latent computation on the transport origin; it does not replace the learned exit-0 readout $q^0=D_\phi(h^0)$. Optional pre-recurrent Transformer layers, when configured, act once at this point rather than inside the recurrent loop. For $t_k=k/K$ and $\Delta t=1/K$, FlowTSFM adds a depth-time embedding and a step-budget embedding, then produces four adaptive-normalization controls:
\begin{equation}
 c_k=e_t(t_k)+e_{\text{depth}}(K),\qquad
 (s_a,c_a,s_m,c_m)=A_\theta(c_k).
 \label{eq:adaln-controls}
\end{equation}
Writing $\operatorname{mod}(z;s,c)=z\odot(1+c)+s$, the attention and MLP sublayers compose a single velocity before the recurrent state is updated:
\begin{align}
 a^k &=\operatorname{MHA}_\theta\!\left(
 \operatorname{mod}(\operatorname{N}_1(h^k);s_a,c_a)\right),\\
 \widetilde h^k&=h^k+a^k,\\
 m^k&=\operatorname{MLP}_\theta\!\left(
 \operatorname{mod}(\operatorname{N}_2(\widetilde h^k);s_m,c_m)\right),\\
 F^k&=a^k+m^k,\qquad
 h^{k+1}=h^k+\Delta t\,F^k.
 \label{eq:detailed-update}
\end{align}
MHA is vanilla non-causal multi-head self-attention with $q/k/v/o$ projections and the patch padding mask. The MLP is $d_{\text{model}}\rightarrow4d_{\text{model}}\rightarrow d_{\text{model}}$ with GeLU. The intermediate sum $\widetilde h^k=h^k+a^k$ is internal to the Transformer parameterization of the field; it is not an Euler transition. The only recurrent transition is the final line of Eq.~\ref{eq:detailed-update}. There are no attention, MLP, or residual gates. All block parameters are reused for every $k$.

\paragraph{Exits and decoder.}
At every selected exit $k\in\mathcal E$, including $k=0$, register positions are removed, final LayerNorm is applied, and the shared output chart $D_\phi$ maps each patch from $d_{\text{model}}$ to $LQ$ values before reshaping to a $T\times Q$ quantile field. Consequently, $q^0=D_\phi(h^0)$ is a learned decoder output; because $\pi^0_{s,j}=r_j$, the $\lambda_0$ term anchors this initial field to the analytic source. In the base model, the input embedding and output chart are residual projection blocks. The Chronos-2-I/O variant replaces those two modules with the exact pretrained-compatible Chronos-2 residual blocks; the FlowTSFM recurrent core and any optional pre-recurrent layers remain separate trainable model parameters.

Table~\ref{tab:hparams} reports the settings of the base accuracy run. The base checkpoint supplies the GIFT-Eval and TIME scores. The Chronos-2-I/O checkpoint supplies the controlled trajectory study because it shares Chronos-2's input/output geometry and the nine common quantile levels used by the matched protocol.

\begin{table}[H]
  \caption{Reported FlowTSFM configuration for the base accuracy run.}
  \label{tab:hparams}
  \centering
  \small
  \setlength{\tabcolsep}{5pt}
  \begin{tabular}{lr}
    \toprule
    Setting & Base accuracy run \\
    \midrule
    Context length / patch size & 8{,}192 / 16 \\
    $d_{\text{model}}$ / attention heads & 1{,}024 / 16\\
    Quantiles / recurrent calls & 99 / 12 \\
    Register tokens & 4 \\
    Parameters & 38{,}845{,}536 \\
    Real / kernel-synthetic data & 50\% / 50\% \\
    Nominal prediction-mask ratio & 40\% \\
    Optimizer   & AdamW \\
   precision  & bfloat16
    \\
    
    Peak / final learning rate & $3\times10^{-4}$ / $10^{-5}$ \\
    Warm-up / weight decay & 5k / 0.1 \\
    $\lambda_{\rm ds},\lambda_{\rm qf}$ & 0.5, 0.3 \\

    \bottomrule
  \end{tabular}
\end{table}

The prior projection is learned jointly with the recurrent block and shared decoder. Its role is to condition the latent starting state; the observable path begins at the learned decoding $q^0$, which Eq.~\ref{eq:objective} anchors to $r$ before supervising the positive-depth intermediate decodings.

\paragraph{Benchmark-score provenance.}
Table~\ref{tab:main-results} reports MASE and CRPS using each benchmark's taskwise Seasonal-Naive normalization followed by geometric aggregation. The FlowTSFM entries are loaded from the completed base evaluations. The Chronos-2 GIFT-Eval entries aggregate its officially submitted results over the 97 configurations shared with the Seasonal-Naive baseline; its TIME entries are the benchmark's reported aggregate over 98 tasks \citep{aksu2024gifteval,ansari2025chronos2,qiao2026time}.

\clearpage

\end{document}